\documentclass[9pt,shortpaper,twoside,web]{ieeecolor}
\usepackage{generic}
\usepackage{cite}
\usepackage{amsmath,amssymb,amsfonts}
\usepackage{algorithmic}
\usepackage{graphicx}
\usepackage{textcomp}

\usepackage{multirow}
\usepackage{tabularx}
\usepackage{caption}
\newcolumntype{L}[1]{>{\raggedright\let\newline\\\arraybackslash\hspace{0pt}}m{#1}}
\newcolumntype{C}[1]{>{\centering\let\newline\\\arraybackslash\hspace{0pt}}m{#1}}
\newcolumntype{R}[1]{>{\raggedleft\let\newline\\\arraybackslash\hspace{0pt}}m{#1}}

\def\BibTeX{{\rm B\kern-.05em{\sc i\kern-.025em b}\kern-.08em
    T\kern-.1667em\lower.7ex\hbox{E}\kern-.125emX}}

\begin{document}
\title{SleepPPG-Net: A deep learning algorithm for robust sleep staging from continuous photoplethysmography}
\author{Kevin Kotzen, Peter H. Charlton, Sharon Salabi, Lea Amar, Amir Landesberg, Joachim A. Behar
\thanks{P. H. Charlton receives funding from the British Heart Foundation grant FS/20/20/34626.}
\thanks{K. Kotzen, A. Landesberg and J. A. Behar are with the Biomedical Engineering Faculty at Technion-Israel Institute of Technology, Haifa, Israel. (e-mail: kevinkotzen@gmail.com, jbehar@technion.ac.il)}
\thanks{P. H. Charlton is with the Department of Public Health and Primary Care at University of Cambridge, Cambridge, UK}
\thanks{P. H. Charlton is with the Research Centre for Biomedical Engineering, University of London, London, UK} 
\thanks{S. Salabi is with the Computer Science Faculty at Technion-Israel Institute of Technology, Haifa, Israel}
\thanks{L. Amar is with Mines ParisTech - PSL University, Paris, France}
}

\maketitle

\begin{abstract}
Sleep staging is an essential component in the diagnosis of sleep disorders and management of sleep health. It is traditionally measured in a clinical setting and requires a labor-intensive labeling process. We hypothesize that it is possible to perform automated robust 4-class sleep staging using the raw photoplethysmography (PPG) time series and modern advances in deep learning (DL). We used two publicly available sleep databases that included raw PPG recordings, totalling 2,374 patients and 23,055 hours. We developed SleepPPG-Net, a DL model for 4-class sleep staging from the raw PPG time series. SleepPPG-Net was trained end-to-end and consists of a residual convolutional network for automatic feature extraction and a temporal convolutional network to capture long-range contextual information. We benchmarked the performance of SleepPPG-Net against models based on the best-reported state-of-the-art (SOTA) algorithms. When benchmarked on a held-out test set, SleepPPG-Net obtained a median Cohen's Kappa ($\kappa$) score of 0.75 against 0.69 for the best SOTA approach. SleepPPG-Net showed good generalization performance to an external database, obtaining a $\kappa$ score of 0.74 after transfer learning. Overall, SleepPPG-Net provides new SOTA performance. In addition, performance is high enough to open the path to the development of wearables that meet the requirements for usage in clinical applications such as the diagnosis and monitoring of obstructive sleep apnea. 
\end{abstract}

\begin{IEEEkeywords}
Sleep staging, Deep Learning, Photoplethysmography, Remote Health
\end{IEEEkeywords}

\section{Introduction}\label{sec:intro}
Sleep is essential for human health, well-being, and longevity \cite{Walker2018}. Insufficient sleep and poor sleep quality are known to cause a myriad of physical and mental diseases such as cardiovascular disease, obesity, and depression \cite{Walker2018}. Sleep disorders such as obstructive sleep apnea (OSA) are highly prevalent, affecting up to one-sixth of the global adult population \cite{Benjafield2019}. Despite the impact on quality of life, many people with sleep disorders are unaware of their condition and remain undiagnosed \cite{Benjafield2019}. 

Sleep disorders are traditionally diagnosed with a sleep study called polysomnography (PSG). During a PSG study, the patient is monitored and observed overnight, usually in a sleep laboratory. The patient is connected to sensors that measure and record several neurophysiological and cardiorespiratory variables \cite{Kapur2017}. PSG data is labeled using the electroencephalogram in a manual or semi-manual manner by a technician trained in sleep scoring. Labels are assigned for each successive 30s window called sleep epochs, henceforth referred to as ``sleep-windows''. The PSG process is uncomfortable for the patient, who has to spend a night in a clinical environment, and labor-intensive, requiring a technician to monitor the patient overnight and another technician to perform manual sleep stage labeling.  Furthermore, the number of clinics that perform PSG are limited and most clinics have long waiting times \cite{Sriram2021}. For example, in Australia and Singapore patients wait an average of 80 days for a PSG examination \cite{Sriram2021, Phua2021}. The limited availability of PSG make repeated studies unfeasible and long-term monitoring of disease progression is currently not an option.

With the recent proliferation of wearable sensors and mobile health applications, there has been a rapid increase in the number of devices that aim to assess sleep quality and disorders more objectively and frequently, particularly targeting the monitoring of the individual in their home environment i.e. outside of the traditional clinical setting \cite{Behar2015, Tan2019, Behar2018, Behar2019, Imtiaz2021}. The accuracy of the sleep metrics obtainable from these wearables is however limited and these devices do not yet meet clinical requirements \cite{Moreno2019, Chinoy2021}. 

Sleep and the autonomous nervous system (ANS) are regulated by the same major central nervous system mechanisms resulting in a strong connection between sleep stage and ANS activity \cite{Fink2018}. The ANS in turn regulates the cardiovascular and respiratory systems which makes these systems a good proxy for sleep measurement \cite{Cabiddu2012}. As reviewed by Ebrahimi et al. \cite{Ebrahimi2021}, research efforts to improve the clinical accuracy of sleep staging from cardiorespiratory waveforms have thus far mostly focused on the development of algorithms that perform sleep staging from the electrocardiogram (ECG). A vast majority of these works used feature engineering (FE) and recurrent neural network (RNN) for automated sleep staging \cite{Fonseca2017, Wei2018, Sun2020, Fonseca2020}. Cohen's Kappa ($\kappa$) performance for this FE-based approach has reached 0.60 \cite{Fonseca2020}. More recently, Sridhar et al. \cite{Sridhar2020} developed a deep learning (DL) model taking as input the instantaneous heart rate (IHR), i.e. a time series derived from the interbeat intervals (IBIs) computed from the ECG. Their DL model consists of a residual convolutional network (ResNet) followed by a temporal convolutional network (TCN). They reported in-domain test $\kappa$ performance of 0.67 for the Sleep Heart Health Study (SHHS) and 0.69 for the Multi-Ethnic Study of Atherosclerosis (MESA), and out-of-domain generalization performance of 0.55 for the PhysioNet/Computing in Cardiology database \cite{Ghassemi2018}.

Most novel wearable sensors are capable of recording continuous photoplethysmography (PPG). There is however significantly less work published on sleep staging from PPG than there is for ECG. Most works that use PPG usually do so in the context of transfer learning (TL), where models are trained on a large database of heart rate variability (HRV) measures and then fine-tuned to a smaller database of pulse rate variability (PRV) measures derived from the IBIs detected on the PPG. These works report $\kappa$ performance approaching 0.66 \cite{Radha2021, Wulterkens2021}. Sleep staging from the raw PPG is a relatively novel approach. In 2020 Korkalainen et al. \cite{Korkalainen2020} used the PPG as input to a convolutional neural network followed by RNN to obtain $\kappa$ performance of 0.55. Most recently Hutten et al. \cite{Huttunen2021}, under the supervision of Korkalainen, updated these results to a $\kappa$ of 0.64. 

This research aims to demonstrate that sleep staging from the raw PPG, using an advanced DL approach, is superior to sleep staging approaches that use features or time series extracted from the IBIs of the PPG. 

\begin{figure*}[h!]
 \centering
 \includegraphics[width=\linewidth]{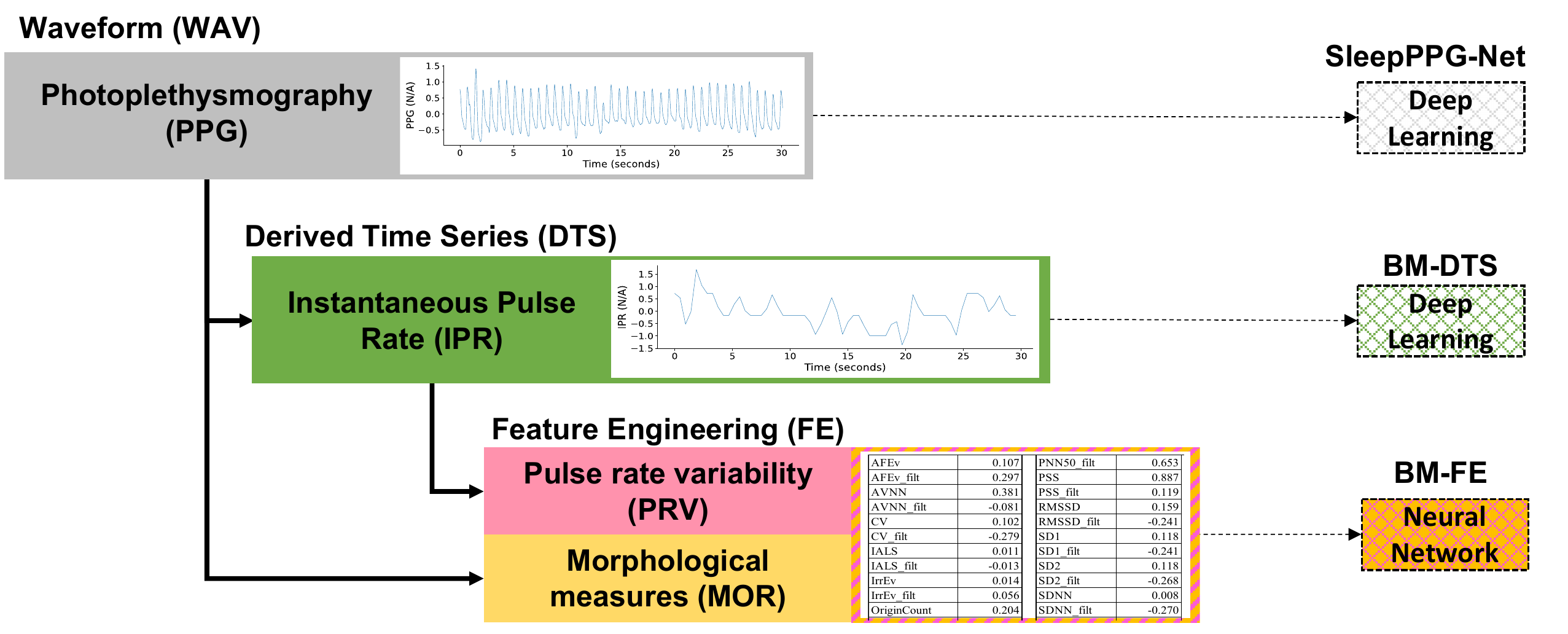}
 \caption[]{Three approaches to ML for sleep staging from PPG. Our new model, SleepPPG-Net takes as input the PPG waveform (WAV).The derived time series (DTS) and feature engineering (FE) approaches allow comparison with SOTA algorithms described in the literature.}
 \label{fig:three_approaches}
\end{figure*}

\section{Methods}\label{sec:methods}
As depicted in Figure \ref{fig:three_approaches}, we considered three ML approaches to sleep staging from PPG. The first approach used handcrafted features engineered from the PPG and a neural network (NN) classifier, the second approach used derived time series (DTS) extracted from IBIs of the PPG as input to a DL classifier, and the third approach used the minimally-preprocessed PPG and DL. Models for the approaches are named Benchmark-FE (BM-FE), Benchmark-DTS (BM-DTS), and our new algorithm SleepPPG-Net. All models used a sequence-to-sequence architecture and were trained end-to-end \cite{Sutskever2014}. 

\subsection{Databases}\label{sec:methods_databases}
Permission to use retrospective medical databases was granted following Technion-IIT Rappaport Faculty of Medicine institutional review board approval 62-2019. We used three labeled PSG databases in our experiments. SHHS Visit 1 \cite{Quan1997, Zhang2018}, totaling 5,758 unique patients, was used for model pretraining. MESA \cite{Chen2015, Zhang2018}, totaling 2,056 unique patients, was used for training and testing, and the Cleveland Family Study (CFS) Visit-5 v1 \cite{Redline1995, Zhang2018}, totaling 324 unique patients, was used to evaluate generalization performance both with and without TL. Patients from MESA were randomly allocated to train and tests sets, stratifying by age, gender, and Apnea-Hypopnea Index (AHI). MESA-train contains 1,850 patients and MESA-test 204 patients. Patients from CFS were allocated into folds to support evaluation with TL. CFS-train consists of 4 overlapping folds with 240 patients each and CFS-test has 4 non-overlapping folds of 80 patients each. Performance was evaluated on MESA-test and CFS-test.  Databases are described in more detail in Table \ref{tab:list_of_database}.

All databases were downloaded from the National Sleep Resource Center \cite{Zhang2018} and came with sleep stage labels that were manually assigned by sleep experts from the full PSG \cite{Redline1995, Quan1997, Chen2015, Zhang2018}. Each PSG was labeled only once \cite{Redline1995, Quan1997, Chen2015, Zhang2018}. PPG sampling rates were 256Hz in MESA, and 128Hz in CFS. SHHS does not contain PPG, but we used ECG from this database to pretrain our models. SHHS ECG sampling rate was 512Hz.  All PSG recordings were padded or truncated to a length of 10 hours. We used all patients in our databases, only removing patients where a valid PPG or ECG waveform could not be loaded. In total, 2 patients were removed from MESA and 4 patients were removed from CFS. In the interest of improving reproducibility and comparability, the list of patients assigned to MESA-test is presented in Supplement \ref{secA1}. 

\begin{table}[h!]
\caption{Description of PSG databases used in the experiments. Statistics are presented with their median and interquartile-range (IQR)}
\begin{center}
\setlength\tabcolsep{0pt}
\begin{tabular*}{\linewidth}{p{0.35\linewidth}p{0.23\linewidth}p{0.23\linewidth}p{0.23\linewidth}}
\hline
~               & SHHS  & MESA  & CFS   \\
\hline
Patients        & 5,767  & 2,054  & 320   \\
Gender (M:F)    & 1:1.1 & 1:1.2 & 1:1.2 \\
Total Windows   & 5.83M & 2.35M & 0.37M \\
Duration (hrs)  & 9 [8-9] & 10 [9-10] & 10 [9-10] \\
Age (yrs)  &           63 [55-72] &           68 [62-76] &           42 [21-54]\\
Wake (\%)  & 27 [19-35] &           37 [30-47] &           34 [27-44]\\
Light (\%) & 44 [36-52] &           43 [36-50] &           39 [29-46]\\
Deep (\%)  & 12 [6-18] &           \:\:5 [1-10] &           12 [7-19]\\
REM (\%)  & 14 [10-17] &           11 [7-14] &           11 [8-14]\\
\hline
\end{tabular*}
\label{tab:list_of_database}
\end{center}
\end{table}

\subsection{Sleep stages}
Modern sleep scoring follows guidelines maintained by the American Academy of Sleep Medicine (AASM) \cite{Berry2017}. AASM sleep stages include wake, rapid eye movement (REM), and three non-rapid eye movement (NREM) stages denoted N1, N2, and N3. In this work we consider 4-class sleep staging with classes as follows: wake, light (N1/N2), deep (N3), and REM. CFS and SHHS were not labeled using AASM but rather an older set of guidelines called Rechtschaffen and Kales (R\&K). The major difference between R\&K and AASM is that R\&K contains an additional NREM stage. R\&K NREM stages are denoted S1, S2, S3, and S4. We assign R\&K labels to our 4-classes as follows: wake, light (S1/S2), deep (S3/S4), and REM. 

\subsection{Data Preparation}
 
\subsubsection{PPG preprocessing}\label{sec:methods_raw}
The PPG was filtered and downsampled to form WAV$_{PPG}$. Low-pass filtering removes high-frequency noise and prevents aliasing during down-sampling. We specifically used a low-pass filter as we wished to keep lower frequency components such as breathing and capillary modulation intact. The filter was built using a zero-phase 8th order low-pass Chebyshev Type II filter with a cutoff frequency of $8Hz$ and a stop-band attenuation of 40dB. The filtered PPG was downsampled to $34.1\overline{3}Hz$ using linear interpolation, reducing the computational and memory requirements for ML. We choose a sampling rate of $34.\overline{3}Hz$ as this resulted in 1024 ($2^{10}$) samples per 30s sleep-window. By using a $2^n$ number we could maintain full temporal alignment of data with sleep-windows during ML pooling operations. WAV$_{PPG}$ was cleaned by clipping values to three standard deviations and then standardized by subtracting the mean and dividing by the standard deviation.

\subsubsection{Feature engineering}
Our FE and DTS approaches rely on robust detection of peaks on the PPG. We used a band-pass filter to remove noise from the PPG that would otherwise affect peak detection. This filtering stage was independent of WAV$_{PPG}$ preprocessing. The band-pass filter was designed to have a minimal impact on the morphology of the PPG. Given that the heart beats in a range of around 40-100bpm (0.66Hz–1.66Hz) and based on a review of the literature around the optimal filtering of PPGs \cite{Hara2017, Park2017, Pollreisz2019, Tanweer2017, Temko2017}, we used a band-pass filter with a pass-band of 0.4-8Hz. The filter was built using a zero-phase 8th order band-pass Chebyshev Type II filter with a pass-band of 0.4-8Hz and stop-band attenuation of 40dB. PPG peaks were detected from the filtered time series using an automatic beat detection algorithm developed by Aboy et. al. \cite{Aboy2005} and implemented in the PulseAnalyse toolbox \cite{Charlton2019}. This PPG peak detector was chosen because it demonstrated the highest peak detection performance when evaluated on PPGs recorded during PSG \cite{Kotzen2021}. For SHHS, the ECG peaks were detected using epltd0 \cite{Hamilton2002} a state-of-the-art ECG peak detection algorithm. 

PRV and HRV measures were extracted using the Python HRV features implemented in  \cite{Chocron2020}. This library calculates 21 HRV measures per set of IBIs. Morphological measures (MOR) were extracted from the time domain, first and second order derivatives, and the frequency domain of the PPG. A total of 41 features were extracted using a MOR toolbox developed within the context of this research. We calculated measures for each sleep-window twice. First only for the current sleep-window and then again with the two preceding and proceeding windows included. We did this because HRV measures should be calculated with a time span of at least two and a half minutes, but sleep-windows are only 30s.

We standardized MOR and PRV features on a per-patient basis. The mean and standard deviation of each feature for each patient were standardized to 0 and 1 respectively. This per-patient standardization acts as a form of personalization and eliminates differences in baseline values between patients. 

\subsubsection{Instantaneous pulse rate}
The IHR and instantaneous pulse rate (IPR) were extracted from the IBIs according to the methods described by Sridhar et al. \cite{Sridhar2020}. The only modification made was that we used a re-sampling rate of $2.1\overline{3}Hz$, as opposed to $2.1\overline{3}Hz$ as this yielded 64 ($2^6$) samples per 30s sleep-window. By using a $2^n$ number we could maintain full temporal alignment of data with sleep-windows during ML pooling operations.

\subsection{Machine learning}\label{sec:methods_models}
We define our problem as follows: given a sequence of $L$ ordered sleep-windows, with input signal $S$ and labels $P$, map the input signal to the labels using network $F$ such that $F(S) \mapsto P$. Sleep-windows are indexed with subscript $l$, where $\{l : 1...L\}$ refers to the $l$th sleep-window in the sequence. In line with other sequence-to-sequence models, we break $F$ into parts, namely, a sleep-window encoder $F_E$, a sequence encoder $F_S$, and a classifier $F_C$. The $F_E$ extracts information from each individual $S_l$, translating the high dimensionality inputs into a lower-dimensional space called an embedding $X_l$ such that $F_E(S)\mapsto X$. $F_S$ then exploits the cyclic and phasic nature of sleep and considers the sequence as a whole, adding contextual information to each $X_l$, by looking at neighboring embeddings ${X_{l-i}...X_{l+i}}$, where $i$ is the receptive-field of $F_S$, resulting in a richer representation $Z_l$ such that $F_S(X)\mapsto Z$. Finally, $F_C$ computes a probability prediction of each sleep-stage at each sleep-window $P_l$ from $Z_l$ such that $F_E(Z)\mapsto P$. 

We further define $M$ as demographic data, $n_{x}$ as the size of $S_l$, $n_{e}$ as the size of $X_l$, $n_{z}$ as the size of $Z_l$, $n_{h}$ as the number of hidden units in a RNN, and $C$ as the number of output classes. For 4-class sleep staging $C=4$. We feed $M$, to each model by concatenating $M$ to each $X_l$. 

\subsubsection{BM-FE model}
BM-FE model architecture is similar to the model developed by Radha et al. \cite{Radha2021}. The input $S$ consists of a sequence of PRV and MOR features. $L=1,200$ and $n_{x}=126$. The $F_E$ consists of a 5-layer time-distributed deep neural network (DNN). Time-distribution applies the same encapsulated layer to each temporal slice. The $F_S$ consists of 2-stacked bidirectional long short-term memory (LSTM) layers and the $F_C$ is a 4-layer time-distributed DNN. Dropout is used in the $F_C$ for regularization. $n_{e}=16$, $n_{h}=128$, and $n_{z}=256$. A full description of the model and hyperparameters are presented in Supplement \ref{secA2}. 

\subsubsection{BM-DTS model}
The BM-DTS model architecture was based on Sridhar et al. \cite{Sridhar2020} with some minor modifications.  The input $S$ is the continuous IPR time series. $F_E$ consists of 3 time-distributed residual convolution (ResConv) blocks followed by a time distributed DNN. Each ResConv has 3 1D-convolutions followed by pooling layer and residual addition. $F_S$ uses 2 stacked TCNs. Each TCN consists of 5 dilated 1D-convolutions followed by residual addition and dropout. $F_C$ is simply a 1D-convolution. $L=1200$, $n_x=256$, $n_e=128$ and $n_z=128$. A full description of the model including parameters is presented in Supplement \ref{secA3}.     

\subsubsection{SleepPPG-Net}
\begin{figure*}[h!]
 \centering
 \includegraphics[width=0.75\textwidth]{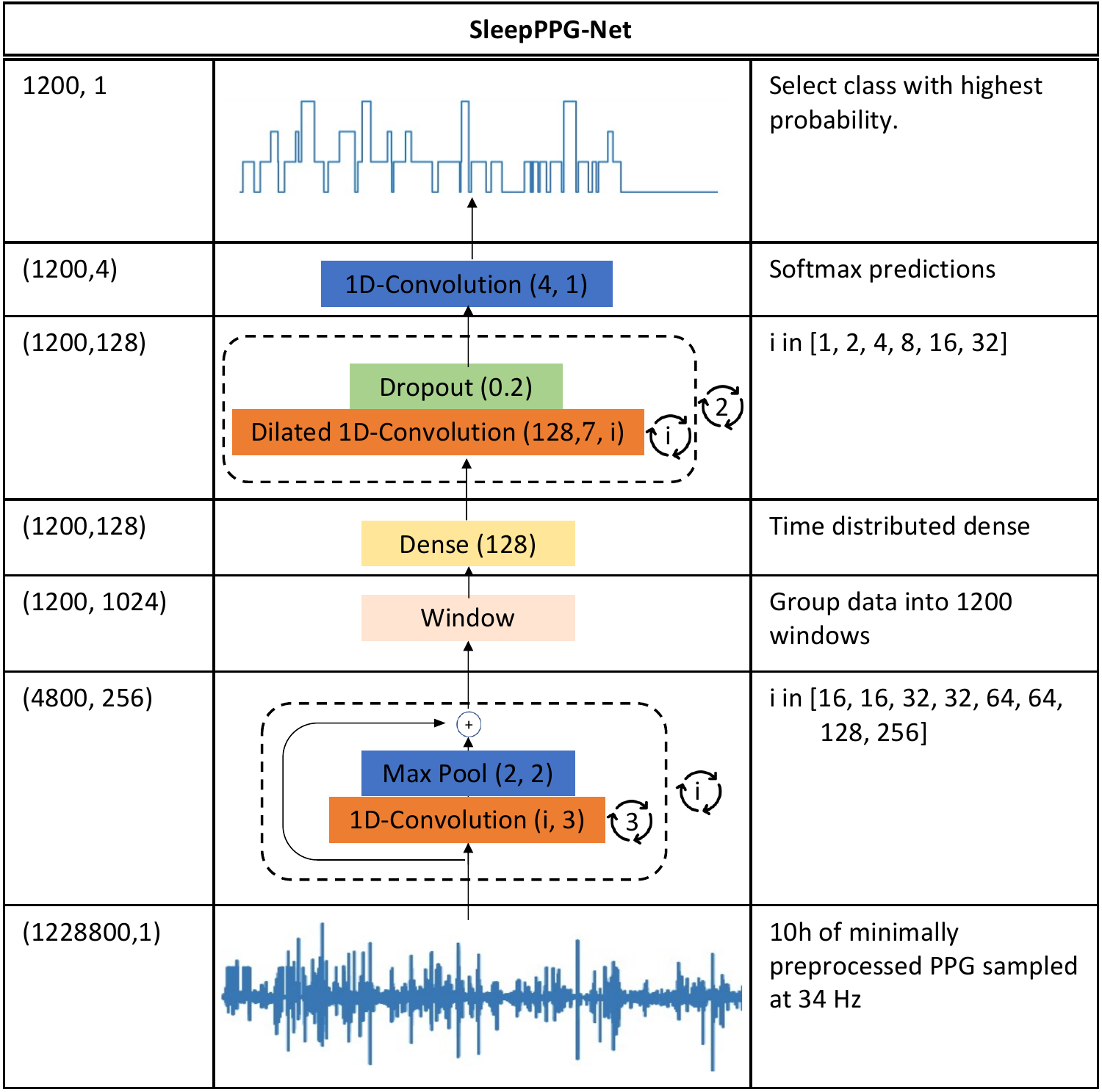}
 \caption[]{Model architecture of SleepPPG-Net, a novel algorithm for 4-class sleep staging from the raw PPG time series. The input to the network is a preprocessed PPG with a sampling rate of 34Hz. An 8-layer ResConv is used to extract increasingly complex features from the input. A TCN with a wide receptive field adds contextual information to the embeddings. This is followed by classification using a 1D-convolution. The classes with the highest probabilities are used to form the predicted hypnogram.}
 \label{fig:SleepPPG-Net-architecture}
\end{figure*}

SleepPPG-Net was inspired by WaveNet \cite{Oord2016}, Wave2Vec \cite{Schneider2019} and Sridhar et al \cite{Sridhar2020}. SleepPPG-Net architecture is shown in Figure \ref{fig:SleepPPG-Net-architecture}. Using $'$ to denote continuous data. The input $S'$ is the continuous WAV$_{PPG}$ time series  with $1,228,800$ samples formed by flattening $L=1200$ and $n_x=1024$. $F_E$ extracts continuous embeddings $X'$ from $S'$. $F_E$ consists of 8 stacked ResConvs. Each ResConv contains 3 1D-convolutions followed by max pooling and residual addition. The ResConvs have a kernel of size 3 and max pooling rate of 2. The number of filters in the ResConvs are 16, 16, 32, 32, 64, 64, 128, 256. The max pooling layer in each ResConv reduces temporal dimensionality by a factor of 2, resulting in $X'$ with a dimension of $4800\times256$. A temporal windowing layer reestablishes $X$ with $L=1200$ by dividing $X'$ into 1200 windows. At this point, $X$ has a dimension of $1200\times1024$. A time-distributed DNN then compresses each $X_l$ so that $n_e=128$. The $F_S$ consists of 2 stacked TCN blocks which add long-range temporal information to $X$ forming $Z$. Each TCN consists of 5 dilated 1D-convolutions followed by residual addition and  dropout. The dilated 1D-convolutions have a kernel size of 7, dilation rates of 1, 2, 4, 8, 16, 32 and filters of size 128 such that $n_z=128$. Finally, $F_C$ uses a 1D-convolution with kernel of size 1 and 4 filters to make predictions $P$ with a shape of $1200\times4$. The Leaky ReLU activation function was used in all layers except the output layer which uses the Softmax activation function.  

\subsubsection{Pretraining}
Pretraining is the process of specifically training a ML model with the intention of using the pretrained model as a starting point for solving other problems. Pretraining improves training convergence times and sometimes improves model performance \cite{Raghu2019}. We pretrained our models on ECG data from the SHHS database. BM-FE was pretrained on HRV measures derived from the ECG. BM-DTS was pretrained on the IHR derived from the IBIs of the ECG and SleepPPG-Net was pretrained on the raw ECG. 

\subsubsection{Transfer learning}
TL is an effective means of adapting a ML model from one domain to another. TL is widely used in medical image analysis, where it has been shown to improve performance when training on small datasets \cite{Raghu2019}. In our work, we use the term $TL$ to denote adaption to a specific external database. We applied TL to our external database using 4-folds. The pretrained model was used as a starting point and each fold was trained and evaluated independently, before all results were brought together and analyzed as a whole. 

\subsubsection{Training}\label{sec:methods_training}
Models were built using Keras 2.6 and trained using a single NVIDIA A100 GPU. Loss was calculated using the categorical cross-entropy loss, and temporal sample weighting was used to address class imbalance and remove padded regions from loss calculations. The Adam optimizer was used. Hyperparameters  for the BM-FE model were selected through manual experimentation. For BM-DTS and SleepPPG-Net we used 100 Bayesian optimization iterations to tune hyperparameters including; the number of ResConv blocks, ResConv kernel size,  ResConv filter size, embedding size, number of TCN blocks, TCN block kernel size and dilation rate, dropout rate, batch size, learning rate and number of training epochs.  Initial weights for convolutional neural network, DNN, and LSTM layers were set with Xavier uniform initialization. 

When training SleepPPG-Net models from scratch, we used a learning rate of $2.5\times10^{-4}$ and trained for 30 epochs. When training models for the With-pretrain and With-TL training schemes we used a learning rate of $1.0\times10^{-4}$ and trained for 5 epochs. A batch size of 8 was used in all experiments. The remaining hyperparameters used are shown in Figure \ref{fig:SleepPPG-Net-architecture}.    

\subsection{Performance measures}\label{sec:methods_metrics}
The models output a probability prediction for each of the 4 sleep stages at each sleep-window in the full sequence. Probabilities were converted into predictions by selecting the class with the highest probability. All padded regions were removed before calculating performance measures. Performance was evaluated using $\kappa$ and $Ac$. We calculate the $\kappa$ and $Ac$ per patient. The hypnogram labels assigned by sleep experts during PSG scoring are considered to be the ground truth and the probabilities obtained from the model are denoted model predictions. The final reported scores represent the median $\kappa$ and median Ac, calculated from all patients in the test set. $Ac$ is the observed agreement over all the examples and is calculated according to Equation \ref{eq:accuracy}. $\kappa$ is calculated according to Equation \ref{eq:kappa} where $Q\equiv Ac$ and $Q_{e}$ is the chance agreement. $L$ is the total number of samples, $C$ is the number of categories, and $n_{cr}$ is the total number of samples of class $c$ counted by rater $r$. Significance of results was computed using the Kolmogorov-Smirnov test for continuous distributions, and the Student's t-test when only the mean and standard deviation were known. 

\begin{equation}
Ac \equiv Q = \frac{P_{correct}}{L}
\label{eq:accuracy}
\end{equation}

\begin{equation}
\kappa = \frac{Q- Q_{e}}{1-Q_{e}} \quad where \quad
Q_{e} = \frac{1}{L^2}\sum_{c=1}^{C} n_{c1}n_{c2}
\label{eq:kappa}
\end{equation}

We evaluated performance across calculated performance metrics per patient population groups to including; age, sex, race, smoking status, apnea severity, hypertension diagnosis, diabetes diagnosis, and beta blocker usage. 

\subsection{Sleep metrics}
Common sleep metrics obtainable from the polysomnography include; total sleep time (TST), sleep efficiency (SE), sleep-stage fractions (FR$_{Light}$, FR$_{Deep}$, FR$_{REM}$), and sleep stage transitions (Transitions). The formulae used to derive each metric are shown in Equations \ref{eq:tst}-\ref{eq:arousals}. We evaluated the degree to which the sleep metrics calculated from the sleep stages predicted by our models matched the ones calculated from the ground truth.  
The degree of agreement was quantified using the mean square error (MSE) and R-Squared errors (R$^2$). 

\begin{equation}
    TST = \sum{Light} + \sum{Deep} + \sum{REM}
\label{eq:tst}
\end{equation}
\begin{equation}
    SE = \frac{TST}{TST+\sum{Wake}} \times 100
\label{eq:se}
\end{equation}
\begin{equation}
    FR_{Stage} = \frac{\sum{Stage}}{TST} \times 100
\label{eq:fractions}
\end{equation}
\begin{equation}
    Transitions =  \sum{Deeper \rightarrow Lighter}
\label{eq:arousals}
\end{equation}

\section{Results}\label{sec:results}
\begin{figure}[ht!]
 \centering
 \includegraphics[width=\linewidth]{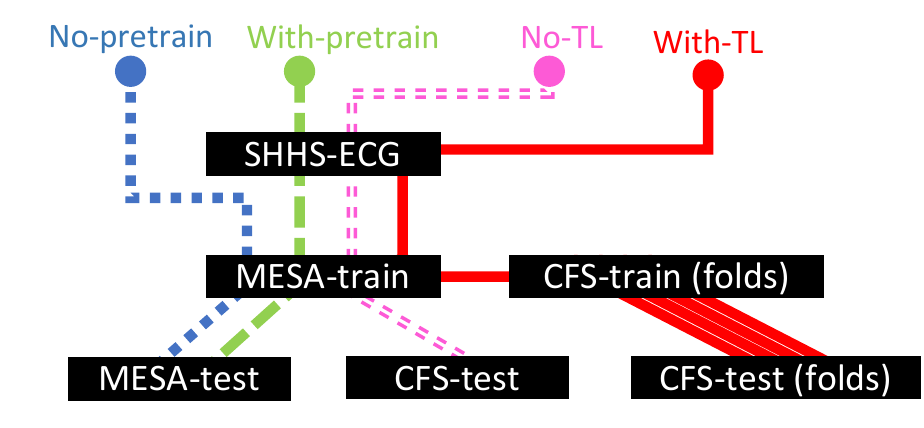}
 \caption[]{Chart depicting the four training schemes used in our experiments.}
 \label{fig:training-schemes}
\end{figure}

We used four different training schemes to train and evaluate the models as depicted in Figure \ref{fig:training-schemes}. Training schemes are: No-pretrain, With-pretrain, No-TL and With-TL. The No-pretrain and With-TL models were trained from Xavier initialization, while the With-pretrain and With-TL models were trained with the weights obtained during pretraining.   

Models are evaluated on a held-out subset of MESA denoted MESA-test, and on 4 non-overlapping folds of the Cleveland Family Study (CFS) denoted CFS-test.  

\subsection{Performance measures}

\begin{table*}[ht!]
\begin{minipage}[t]{0.45\textwidth}\centering
\captionof{table}{Results for models evaluated on MESA-test (n=204). When performed, pretraining used ECG from SHHS.}
\label{tab:baseline_results}
\setlength\tabcolsep{0pt}
\begin{tabular*}{\linewidth}{p{0.32\linewidth}C{0.17\linewidth}C{0.17\linewidth}C{0.17\linewidth}C{0.17\linewidth}}
\hline
~ & \multicolumn{2}{c}{No-pretrain} & \multicolumn{2}{c}{With-pretrain}\\
\hline
Model  & $\kappa$ & $Ac$ & $\kappa$ & Ac\\
\hline
BM-FE        & 0.66 & 78 & 0.66   & 78 \\
BM-DTS       & 0.64  & 76 & 0.69  & 80 \\
SleepPPG-Net & 0.74 & 83 & \textbf{0.75}  & \textbf{84} \\
\hline
\end{tabular*}

\vspace{0.25cm}

\begin{center}
\includegraphics[width=\linewidth]{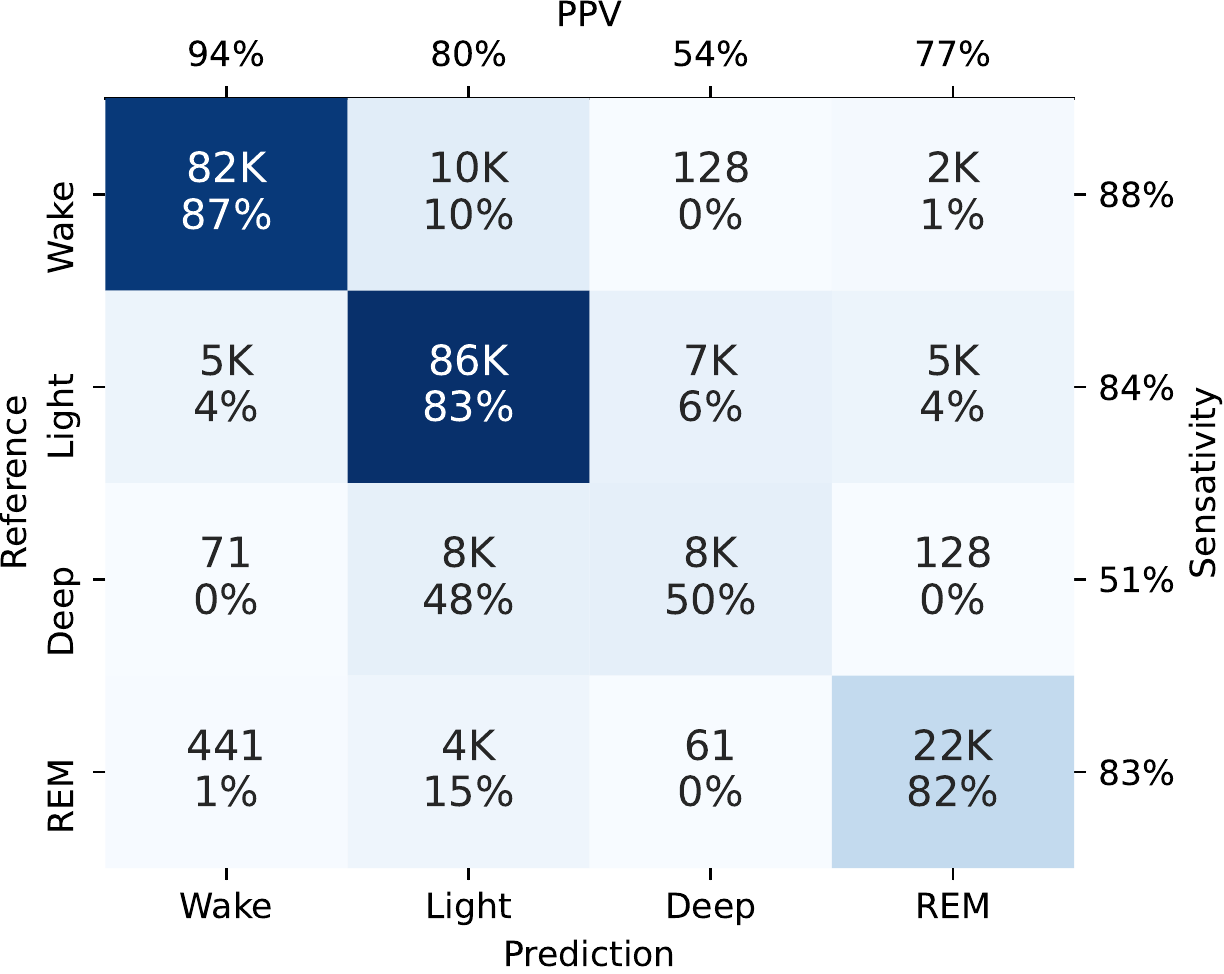}
\captionof{figure}{Confusion matrix for pretrained SleepPPG-Net, evaluated on MESA-test (n=204).}
\label{fig:confusion_matrix_pretrain}
\end{center}

\vspace{0.25cm}

\begin{center}
\includegraphics[width=\linewidth]{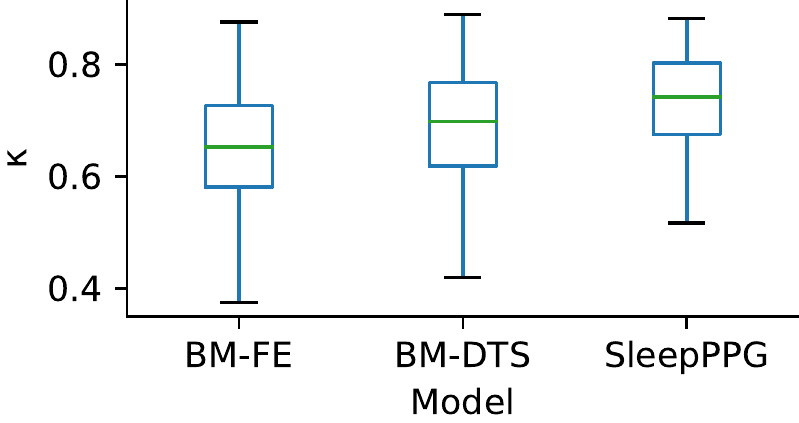}
\captionof{figure}{Distribution of $\kappa$ performance for BM-FE, BM-DTS and SleepPPG on MESA-test (n=204). All models were pretrained on ECG from SHHS and then trained on MESA-train.}
\label{fig:compare_models_boxplot}
\end{center}

\end{minipage}
\hspace{0.5cm}
\begin{minipage}[t]{0.45\textwidth}
\captionof{table}{Results for models evaluated on CFS-test (n=320). When performed, TL used 4 non-overlapping folds.}
\label{tab:generalization_results}
\setlength\tabcolsep{0pt}
\begin{tabular*}{\linewidth}{p{0.32\linewidth}C{0.17\linewidth}C{0.17\linewidth}C{0.17\linewidth}C{0.17\linewidth}}
\hline
~ & \multicolumn{2}{c}{No-TL} & \multicolumn{2}{c}{With-TL}\\
\hline
Model  & $\kappa$ & $Ac$ & $\kappa$ & Ac\\
\hline
BM-FE        & 0.47 & 63 & 0.64 & 76 \\
BM-DTS       & 0.53  & 69 & 0.70  & 79 \\
SleepPPG-Net & 0.67 & 76 & \textbf{0.74}  & \textbf{82} \\
\hline
\end{tabular*}
\vspace{0.25cm}
\begin{center}
\includegraphics[width=\linewidth]{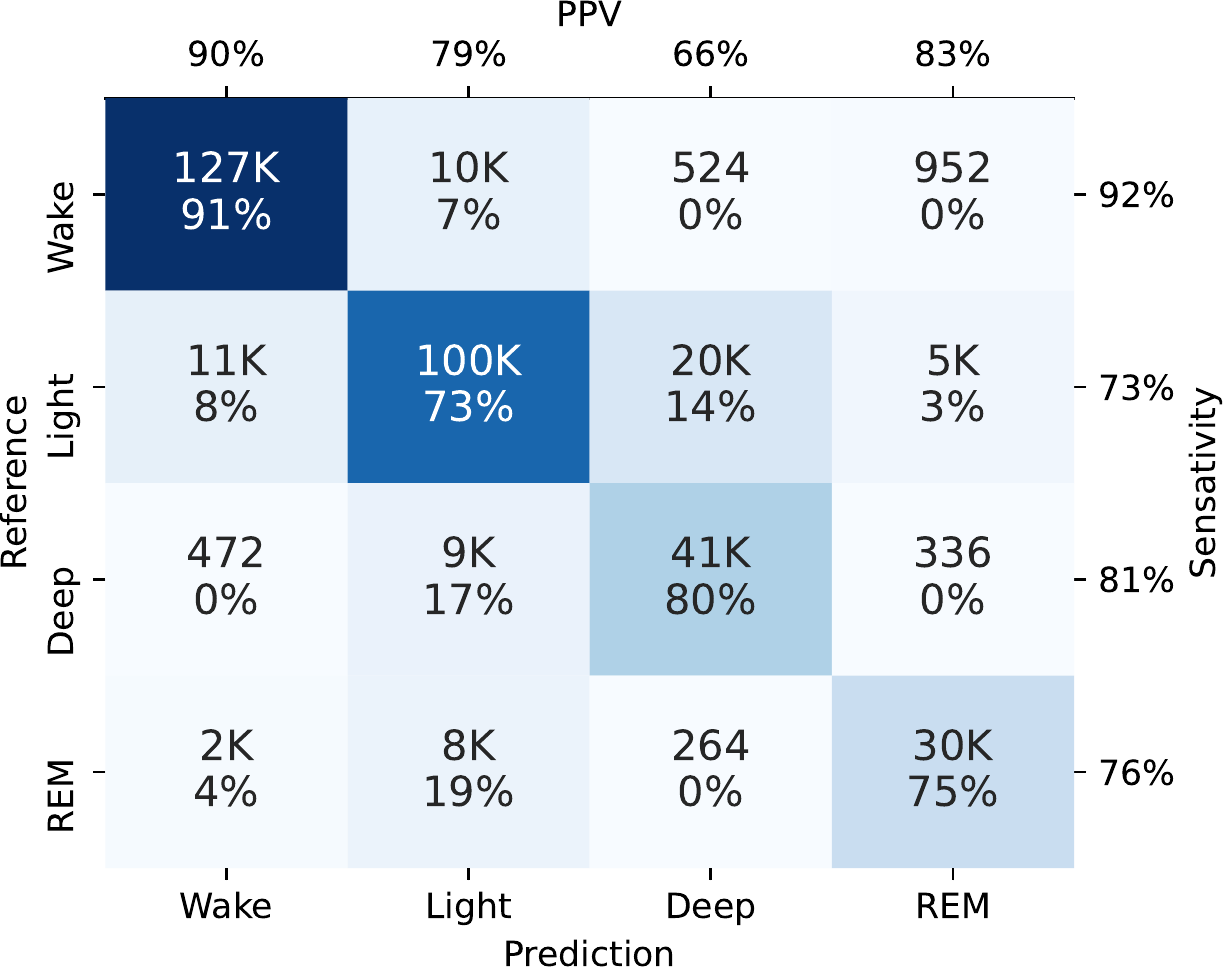}\captionof{figure}{Confusion matrix for SleepPPG-Net with TL, evaluated on CFS-test (n=320).\\}
 \label{fig:confusion_matrix_TL}
 \end{center}
\begin{center}
\includegraphics[width=\linewidth]{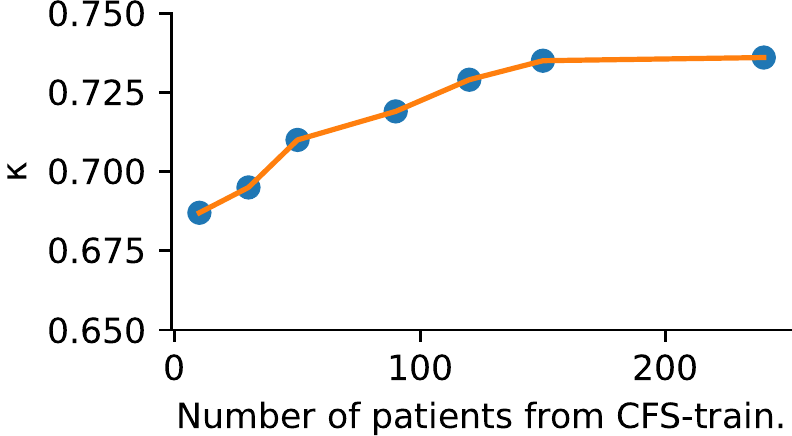}
\captionof{figure}{TL performance on CFS-test by number of patients from CFS-train used for TL. A random subset of patients is taken from each CFS-train fold for training. Experiments were run multiple times and the average is shown.}
\label{fig:transfer-learning-patients}
\end{center}

\end{minipage}
\end{table*}

Table \ref{tab:baseline_results} presents evaluation results for MESA-test. We show performance for models trained with Xavier initialization and on MESA-train (i.e. No-pretrain) and then for models pretrained on ECG from SHHS (i.e. With-pretrain).
Pretraining on SHHS did not have an important effect on the performance of BM-FE or SleepPPG-Net models but significantly improved ($p=0.0002$, Kolmogorov-Smirnov test) BM-DTS performance from a $\kappa$ of 0.64 (0.56 to 0.72) and accuracy ($Ac$) of 76\% to a $\kappa$ of 0.69 (0.62 to 0.77) and $Ac$ of 80\%. The best performing model was the pretrained SleepPPG-Net which scored a $\kappa$ of 0.75 (0.69 to 0.81) and $Ac$ of 84\%. The confusion matrix for the pretrained SleepPPG-Net is presented in Figure \ref{fig:confusion_matrix_pretrain}. The $\kappa$ distribution for the pretrained BM-FE, BM-DTS, and SleepPPG-Net models are compared in Figure \ref{fig:compare_models_boxplot}. 

Table \ref{tab:generalization_results} presents generalization to external test set results using CFS-test for evaluation. We first show performance for models with pretraining on ECG from SHHS and MESA-train but before TL (i.e. No-TL), and then show performance for the same models after applying TL using CFS-train (i.e. With-TL). 
Before TL SleepPPG-Net scored a $\kappa$ 0.67 (0.55 to 0.74) and $Ac$ of 76\%. With TL SleepPPG-Net scored a $\kappa$ of 0.74 (0.66 to 0.79) and $Ac$ of 82\% which is significantly better ($p=0.0005$, Kolmogorov-Smirnov test). The confusion matrix of SleepPPG-Net with TL is presented in Figure \ref{fig:confusion_matrix_TL}. To determine the number of patients needed for effective TL, we evaluate the $\kappa$ performance on CFS-test as a function of the number of patients used for TL as depicted in Figure \ref{fig:transfer-learning-patients}. Performance for SleepPPG-Net with TL improved from a $\kappa$ of 0.68 (0.56 to 0.75) when using only 10 patients to 0.73 (0.64 to 0.78) when using 120 patients. 

\begin{figure*}[h!]
 \centering
 \includegraphics[width=0.98\textwidth]{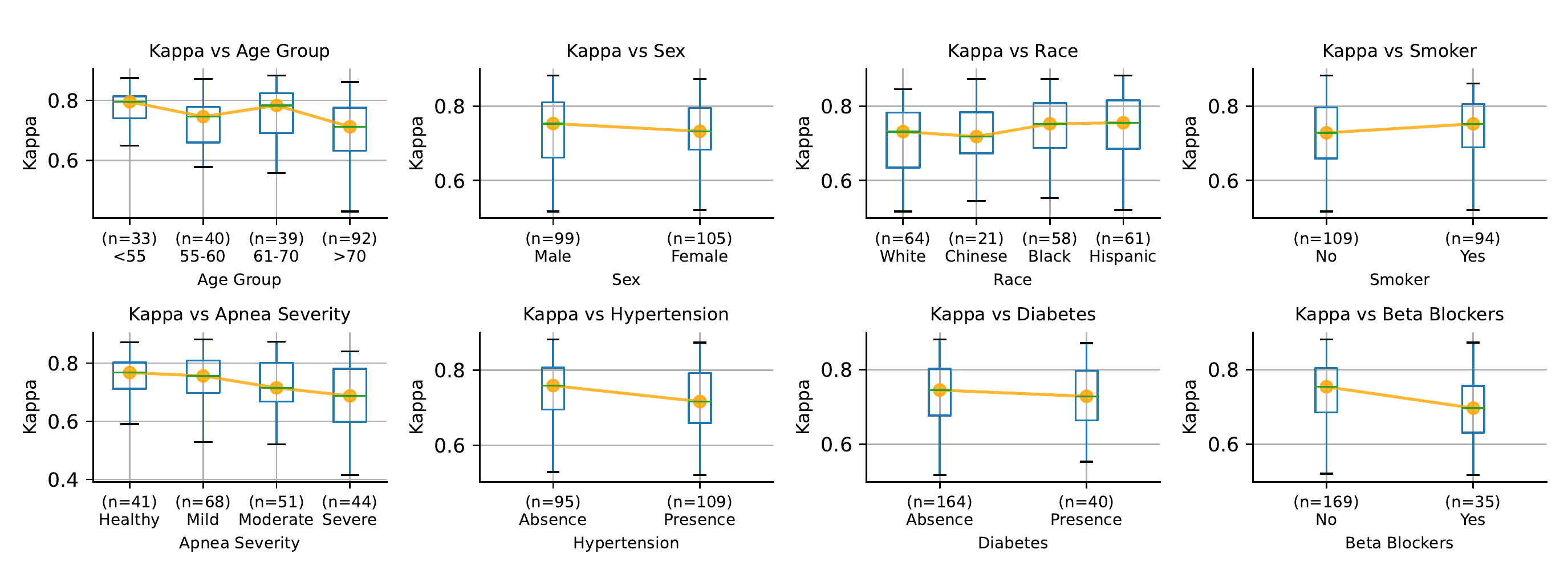}
 \caption[]{SleepPPG-Net With-pretrain performance per clinical group evaluated on MESA-test (n=204).}
 \label{fig:kappa-per-group}
\end{figure*}

The per group $\kappa$ performance is presented in Figure \ref{fig:kappa-per-group}.  Performance is not affected by gender, race, or presence of diabetes. Performance is lower in patient groups with higher apnea severity, older age, hypertension diagnosis and beta blocker usage. Performance is higher for patients that smoke.

\subsection{Sleep metrics}
\begin{figure}
\begin{center}
 \includegraphics[width=\linewidth]{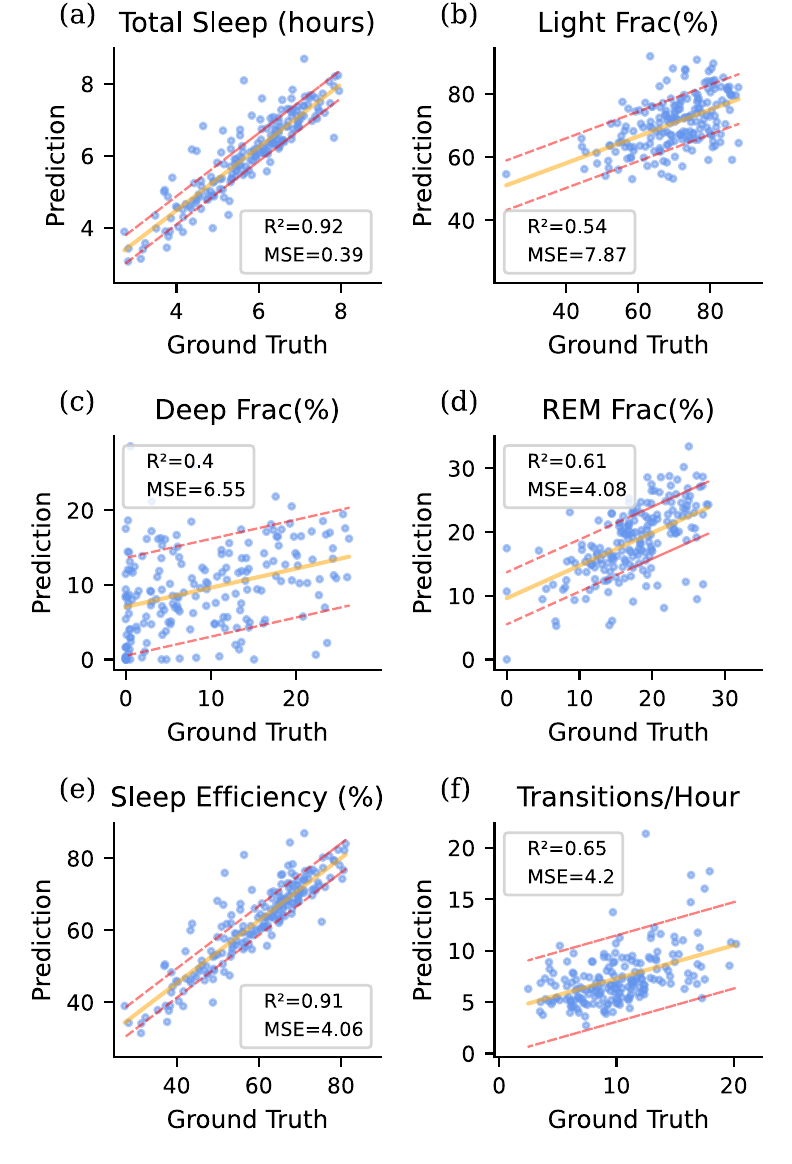}
  \end{center}
 \caption{Comparison between sleep metrics of MESA-test (n=204) calculated from pretrained SleepPPG-Net to those calculated using ground truth. Dotted line shows the MSE.}
\label{fig:predicted_vs_model}
\end{figure}

We evaluated sleep metrics for MESA-test (n=204) using the pretrained SleepPPG-Net. In Figure \ref{fig:predicted_vs_model} we compare the predicted sleep metrics to those calculated from the ground truth. The pretrained SleepPPG-Net scored a MSE of 0.39 hours for total sleep, 7.87\% for Light fraction, 6.55\% for Deep fraction, 4.08\% for REM fraction, 4.1\% for Sleep Efficiency, and 4.2 transitions/hour for Transitions.   

\begin{figure*}[h!]
 \centering
 \includegraphics[width=\textwidth]{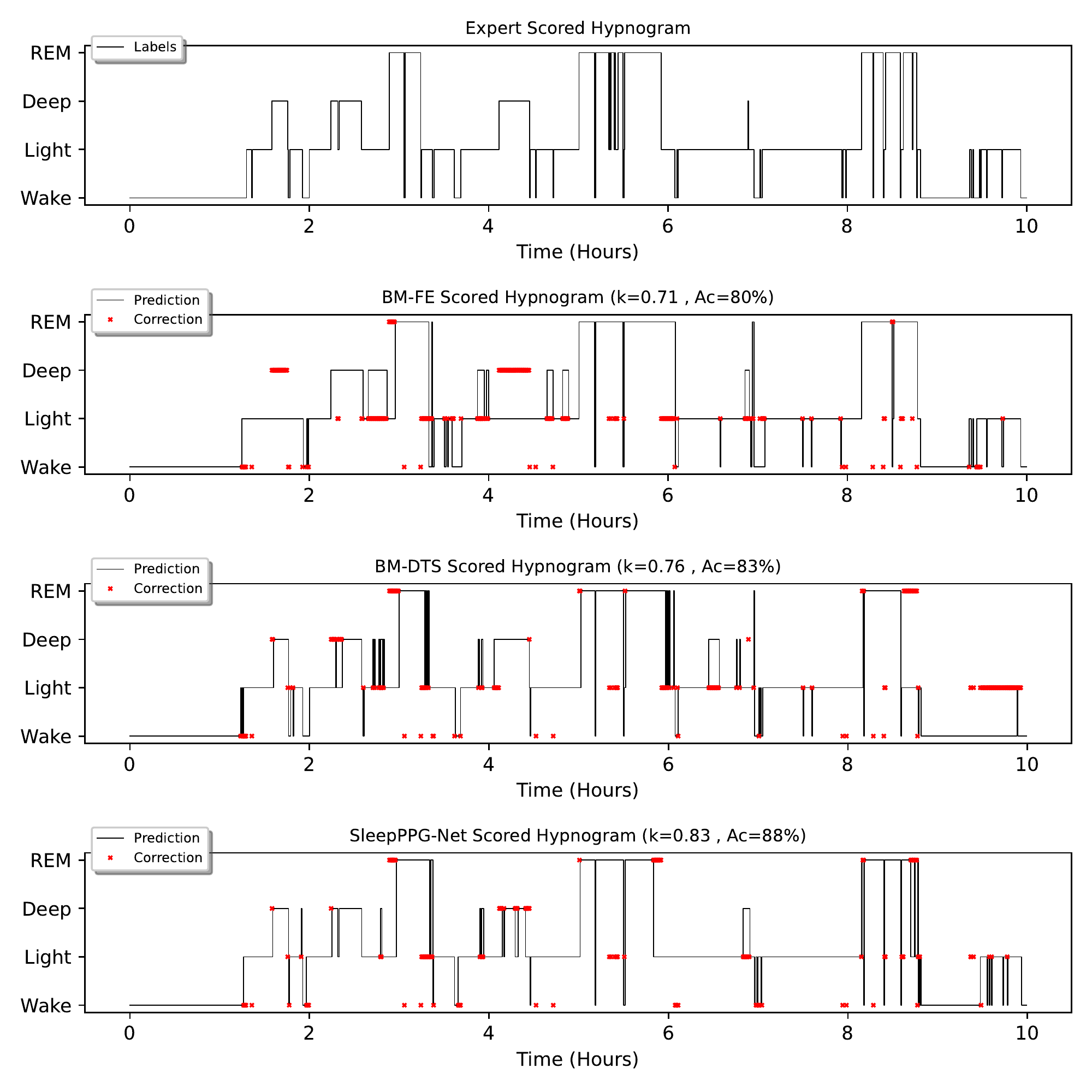}
 \caption[]{Hypnogram showing discrepancies between sleep stage predictions and the hypnogram labels assigned by human scorers during PSG. Hypnogram shown is from MESA patient (ID-0310) who is in MESA-test. }
 \label{fig:hypnograms}
\end{figure*}

\section{Discussion}\label{sec:discussion}
When interpreting the performance of automated sleep staging algorithms it is important to keep in mind that manual scoring by humans is highly subjective \cite{Huy2021}. Inter-rater agreement for PSG labeled by human scorers is reported as a $\kappa$ of 0.76  (95\% confidence interval, 0.71–0.81) \cite{Lee2021} for 5-class sleep staging. Common mistakes between human scorers during PSG include confusion between wake and light sleep and light sleep and deep sleep \cite{Danker-Hopfe2009}. While our problem is somewhat simplified in that we consider 4-class sleep staging, these values provide a sense of the highest performance that may be reached by data-driven algorithms. 

The first important contribution of this research is the novel SleepPPG-Net algorithm. SleepPPG-Net was demonstrated to significantly ($p < 0.0004$, Kolmogorov-Smirnov test) outperform SOTA algorithms including BM-FE and BM-DTS. On the held out test set of 204 MESA patients, SleepPPG-Net scored a $\kappa$ of 0.75 against 0.66 for BM-FE and 0.69 for BM-DTS approaches. SleepPPG-Net performance is also significantly ($p<0.001$, two-sample t-test) higher than the current published SOTA results for sleep staging from PPG which stand at a $\kappa$ of 0.66 \cite{Radha2021, Wulterkens2021}, and significantly ($p=0.02$, two-sample t-test) higher than the current SOTA results for sleep-sleep staging from ECG which are reported at $\kappa$ of 0.69 \cite{Sridhar2020}. Figure \ref{fig:hypnograms} presents an example of the hypnograms generated by BM-FE, DB-DTS and SleepPPG-Net for a single patient. Performance for this patient is best for the SleepPPG-Net model which accurately detects all important sleep structures. We believe that the improved performance achieved by SleepPPG-Net over other approaches can be attributed to several factors. 
First, SleepPPG-Net does not require the annotation of fiduciaries using a PPG peak detector. PPG peak detectors are sensitive to noise and are often unable to handle irregular health rhythms. This may result in noisy and inaccurate IBIs which are relied upon by FE and DTS approaches. 
Second, SleepPPG-Net extracts relevant features from the data automatically thus going beyond domain knowledge . FE approaches use PRV and MOR measures which have been developed as measures of general cardiovascular functioning and may not be optimized for sleep staging.
Third, in using only IBI data, any information contained within the PPG that is not directly related to the heart rate is lost. We included MOR measures in an attempt to include some of this information in our BM-FE model, but as previously stated, these measures are not optimized to sleep staging. Additional information embedded in the raw PPG may include respiration rate, blood pressure, stroke volume, cardiac contractility, peripheral vascular tone and pulse-transit time which are all regulated by the ANS \cite{Elgendi2012, Li2018}.
Finally, the choice of sequence encoder used in SleepPPG-Net is important. The TCN is likely better suited to extract the long-term contextual information than the RNN used in the BM-FE model. Similar performance was maintained for all clinical groups with no important outliers. The small differences that are observed between groups are likely due to the model's inability to accurately distinguish between light and deep sleep and detect short awakenings.  

The second important finding of the research is that pretraining SleepPPG-Net on a PSG database with ECG (thus pretraining in another domain) proved to be an effective means of speeding up training convergence. When trained from scratch, SleepPPG-Net needs to be trained for 30 epochs, whereas when trained from the ECG domain pretrained model, convergence was reached after only 5 epochs. Given the ease that SleepPPG-Net adapts to PPG from ECG, we expect that our pretrained SleepPPG-Net model can be leveraged to develop models with new signal-domains such as wrist-based PPG used in wearables.  

The third important finding of this research is that SleepPPG-Net demonstrates good generalizability, scoring a $\kappa$ of 0.67 (0.55 to 0.74) on CFS-test with no TL step. This is markedly higher than the generalization performance reported by Sridhar et al. \cite{Sridhar2020}, whose model scored a $\kappa$ of 0.55 on the PhysioNet/Computing in Cardiology Sleep database \cite{Ghassemi2018}. With TL, performance of SleepPPG-Net increased significantly reaching a $\kappa$ of 0.74 (0.66 to 0.79). The number of CFS patients needed for effective adaption to CFS is shown to be 120. However, even with only 50 patients performance reaches a $\kappa$ of 0.71 (0.63 to 0.79). These results are promising as they indicate that SleepPPG-Net can effectively be fine-tuned to a new population sample using significantly fewer patients than was required for its original training. This will reduce the time and cost involved in the development of new sleep staging devices.

\textbf{Recommendation:} We recommend that sleep staging from PPG be performed using the raw PPG time series and SleepPPG-Net architecture. To obtain optimal results we suggest pretraining SleepPPG-Net with ECG from a large sleep databases such as SHHS before training on PPG from MESA or another sleep dataset. For optimal generalization performance to a new database, transfer learning with at least 120 patients should be used.

\textbf{Limitations:}
An analysis of per class performance shows that SleepPPG-Net struggles in some areas. Deep sleep is consistently underestimated and is often confused with light sleep. This is likely due to the similarity of the cardiovascular and pulmonary characteristics expressed during deep and light sleep. For applications such as the general detection of OSA, this may not be a problem as light and deep sleep can be grouped without affecting diagnosis \cite{LeeChi2015}. However for disorders such as night terrors or sleepwalking and it is important to distinguish between light and deep sleep \cite{LeeChi2015}. The detection of sleep fragmentation is another issue. Our model fails to reliably detect very short awakenings. It is possible that while these changes are visible in the EEG, they are too rapid to be reflected by the cardiovascular activity. Wake periods that are longer than 1.5 minutes are accurately detected.  While the PPG is sensitive to movement, the incorporation of accelerometer data is likely to provide even better performance. Most wearables already contain an accelerometer which is used for activity tracking.

\section{Conclusion}
SleepPPG-Net demonstrates SOTA performance for sleep staging from PPG. SleepPPG-Net is shown to perform well across patient groups and is easily adapted to new databases and measurement settings. As such, SleepPPG-Net paves the way for the development of sleep staging applications from wearable that are accurate enough for clinical diagnosis. This will allow for improved detection, monitoring, and treatment of sleep disorders in the general population.   

\bibliographystyle{IEEEtran}
\bibliography{generic-color-brief}

\begin{thebibliography}{10}
\providecommand{\url}[1]{#1}
\csname url@samestyle\endcsname
\providecommand{\newblock}{\relax}
\providecommand{\bibinfo}[2]{#2}
\providecommand{\BIBentrySTDinterwordspacing}{\spaceskip=0pt\relax}
\providecommand{\BIBentryALTinterwordstretchfactor}{4}
\providecommand{\BIBentryALTinterwordspacing}{\spaceskip=\fontdimen2\font plus
\BIBentryALTinterwordstretchfactor\fontdimen3\font minus
  \fontdimen4\font\relax}
\providecommand{\BIBforeignlanguage}[2]{{%
\expandafter\ifx\csname l@#1\endcsname\relax
\typeout{** WARNING: IEEEtran.bst: No hyphenation pattern has been}%
\typeout{** loaded for the language `#1'. Using the pattern for}%
\typeout{** the default language instead.}%
\else
\language=\csname l@#1\endcsname
\fi
#2}}
\providecommand{\BIBdecl}{\relax}
\BIBdecl

\bibitem{Walker2018}
M.~Walker, \emph{{Why we sleep}}.\hskip 1em plus 0.5em minus 0.4em\relax
  London: Penguin Books, 2018.

\bibitem{Benjafield2019}
A.~V. Benjafield, N.~T. Ayas, P.~R. Eastwood, R.~Heinzer, M.~S. Ip, M.~J.
  Morrell, C.~M. Nunez, S.~R. Patel, T.~Penzel, J.~L.~D. P{\'{e}}pin, P.~E.
  Peppard, S.~Sinha, S.~Tufik, K.~Valentine, and A.~Malhotra, ``{Estimation of
  the global prevalence and burden of obstructive sleep apnoea: a
  literature-based analysis},'' \emph{The Lancet Respiratory Medicine}, vol.~7,
  no.~8, pp. 687--698, 8 2019.

\bibitem{Kapur2017}
V.~K. Kapur, D.~H. Auckley, S.~Chowdhuri, D.~C. Kuhlmann, R.~Mehra, K.~Ramar,
  and C.~G. Harrod, ``{Clinical Practice Guideline for Diagnostic Testing for
  Adult Obstructive Sleep Apnea: An American Academy of Sleep Medicine Clinical
  Practice Guideline},'' \emph{Journal of Clinical Sleep Medicine}, vol.~13,
  no.~3, pp. 479--504, 2017.

\bibitem{Sriram2021}
B.~Sriram, V.~Singh, S.~Bandaralage, and J.~Bashford, ``{P135 An audit
  investigating the length of time from GP referral to diagnostic
  polysomnography testing in an Australian tertiary center},'' \emph{SLEEP
  Advances}, vol.~2, no. Supplement{\_}1, pp. A65--A65, 10 2021.

\bibitem{Phua2021}
C.~Q. Phua, I.~J. Jang, K.~B. Tan, Y.~Hao, S.~R.~B. Senin, P.~R. Song, R.~Y.
  Soh, and S.~T. Toh, ``{Reducing cost and time to diagnosis and treatment of
  obstructive sleep apnea using ambulatory sleep study: a Singapore sleep
  centre experience},'' \emph{Sleep and Breathing}, vol.~25, no.~1, pp.
  281--288, 3 2021.

\bibitem{Behar2015}
J.~Behar, A.~Roebuck, M.~Shahid, J.~Daly, A.~Hallack, N.~Palmius, J.~Stradling,
  and G.~D. Clifford, ``{SleepAp: An Automated Obstructive Sleep Apnoea
  Screening Application for Smartphones},'' \emph{IEEE Journal of Biomedical
  and Health Informatics}, vol.~19, no.~1, pp. 325--331, 1 2015.

\bibitem{Tan2019}
X.~Tan, J.~D. Cook, J.~Cedernaes, and C.~Benedict, ``{Consumer sleep trackers:
  a new tool to fight the hidden epidemic of obstructive sleep apnoea?}''
  \emph{The Lancet Respiratory Medicine}, vol.~7, no.~12, p. 1012, 12 2019.

\bibitem{Behar2018}
J.~A. Behar, A.~A. Rosenberg, I.~Weiser-Bitoun, O.~Shemla, A.~Alexandrovich,
  E.~Konyukhov, and Y.~Yaniv, ``{PhysioZoo: a novel open access platform for
  heart rate variability analysis of mammalian electrocardiographic data},''
  \emph{Frontiers in physiology}, vol.~9, p. 1390, 2018.

\bibitem{Behar2019}
J.~A. Behar, N.~Palmius, Q.~Li, S.~Garbuio, F.~P. Rizzatti, L.~Bittencourt,
  S.~Tufik, and G.~D. Clifford, ``{Feasibility of Single Channel Oximetry for
  Mass Screening of Obstructive Sleep Apnea},'' \emph{EClinicalMedicine},
  vol.~11, pp. 81--88, 5 2019.

\bibitem{Imtiaz2021}
S.~A. Imtiaz, ``{A Systematic Review of Sensing Technologies for Wearable Sleep
  Staging},'' \emph{Sensors 2021, Vol. 21, Page 1562}, vol.~21, no.~5, p. 1562,
  2 2021.

\bibitem{Moreno2019}
F.~Moreno-Pino, A.~Porras-Segovia, P.~L{\'{o}}pez-Esteban, A.~Art{\'{e}}s, and
  E.~Baca-Garc{\'{i}}a, ``{Validation of Fitbit Charge 2 and Fitbit Alta HR
  Against Polysomnography for Assessing Sleep in Adults With Obstructive Sleep
  Apnea},'' \emph{Journal of Clinical Sleep Medicine}, vol.~15, no.~11, pp.
  1645--1653, 11 2019.

\bibitem{Chinoy2021}
E.~D. Chinoy, J.~A. Cuellar, K.~E. Huwa, J.~T. Jameson, C.~H. Watson, S.~C.
  Bessman, D.~A. Hirsch, A.~D. Cooper, S.~P. Drummond, and R.~R. Markwald,
  ``{Performance of seven consumer sleep-tracking devices compared with
  polysomnography},'' \emph{Sleep}, vol.~44, no.~5, 5 2021.

\bibitem{Fink2018}
A.~M. Fink, U.~G. Bronas, and M.~W. Calik, ``{Autonomic regulation during sleep
  and wakefulness: a review with implications for defining the pathophysiology
  of neurological disorders},'' \emph{Clinical autonomic research : official
  journal of the Clinical Autonomic Research Society}, vol.~28, no.~6, p. 509,
  12 2018.

\bibitem{Cabiddu2012}
R.~Cabiddu, S.~Cerutti, G.~Viardot, S.~Werner, and A.~M. Bianchi, ``{Modulation
  of the sympatho-vagal balance during sleep: Frequency domain study of heart
  rate variability and respiration},'' \emph{Frontiers in Physiology}, vol. 3
  MAR, p.~45, 2012.

\bibitem{Ebrahimi2021}
F.~Ebrahimi and I.~Alizadeh, ``{Automatic sleep staging by cardiorespiratory
  signals: a systematic review},'' \emph{Sleep and Breathing}, 2021.

\bibitem{Fonseca2017}
``{Validation of Photoplethysmography-Based Sleep Staging Compared With
  Polysomnography in Healthy Middle-Aged Adults},'' \emph{Sleep}, vol.~40,
  no.~7, 7 2017.

\bibitem{Wei2018}
R.~Wei, X.~Zhang, J.~Wang, and X.~Dang, ``{The research of sleep staging based
  on single-lead electrocardiogram and deep neural network},'' \emph{Biomedical
  Engineering Letters}, vol.~8, no.~1, p.~87, 2 2018.

\bibitem{Sun2020}
H.~Sun, W.~Ganglberger, E.~Panneerselvam, M.~J. Leone, S.~A. Quadri,
  B.~Goparaju, R.~A. Tesh, O.~Akeju, R.~J. Thomas, and M.~B. Westover, ``{Sleep
  staging from electrocardiography and respiration with deep learning},''
  \emph{Sleep}, vol.~43, no.~7, 2020.

\bibitem{Fonseca2020}
P.~Fonseca, M.~M. van Gilst, M.~Radha, M.~Ross, A.~Moreau, A.~Cerny,
  P.~Anderer, X.~Long, J.~P. van Dijk, and S.~Overeem, ``{Automatic sleep
  staging using heart rate variability, body movements, and recurrent neural
  networks in a sleep disordered population},'' \emph{Sleep}, vol.~43, no.~9,
  pp. 1--10, 2020.

\bibitem{Sridhar2020}
N.~Sridhar, A.~Shoeb, P.~Stephens, A.~Kharbouch, D.~B. Shimol, J.~Burkart,
  A.~Ghoreyshi, and L.~Myers, ``{Deep learning for automated sleep staging
  using instantaneous heart rate},'' \emph{npj Digital Medicine}, vol.~3,
  no.~1, 2020.

\bibitem{Ghassemi2018}
M.~M. Ghassemi, B.~E. Moody, L.-W.~H. Lehman, C.~Song, Q.~Li, H.~Sun, R.~G.
  Mark, M.~B. Westover, and G.~D. Clifford, ``{You Snooze, You Win: the
  PhysioNet/Computing in Cardiology Challenge 2018}.''

\bibitem{Radha2021}
M.~Radha, P.~Fonseca, A.~Moreau, M.~Ross, A.~Cerny, P.~Anderer, X.~Long, and
  R.~M. Aarts, ``{A deep transfer learning approach for wearable sleep stage
  classification with photoplethysmography},'' \emph{npj Digital Medicine},
  vol.~4, no.~1, p. 135, 12 2021.

\bibitem{Wulterkens2021}
B.~M. Wulterkens, P.~Fonseca, L.~W.~A. Hermans, M.~Ross, A.~Cerny, P.~Anderer,
  X.~Long, J.~P.~v. Dijk, N.~Vandenbussche, S.~Pillen, M.~M.~v. Gilst, and
  S.~Overeem, ``{It is All in the Wrist: Wearable Sleep Staging in a Clinical
  Population versus Reference Polysomnography},'' \emph{Nature and Science of
  Sleep}, vol.~13, p. 885, 2021.

\bibitem{Korkalainen2020}
H.~Korkalainen, J.~Aakko, B.~Duce, S.~Kainulainen, A.~Leino, S.~Nikkonen, I.~O.
  Afara, S.~Myllymaa, J.~T{\"{o}}yr{\"{a}}s, and T.~Lepp{\"{a}}nen, ``{Deep
  learning enables sleep staging from photoplethysmogram for patients with
  suspected sleep apnea},'' \emph{Sleep}, no. May, pp. 1--10, 5 2020.

\bibitem{Huttunen2021}
R.~Huttunen, T.~Lepp{\"{a}}nen, B.~Duce, A.~Oksenberg, S.~Myllymaa,
  J.~T{\"{o}}yr{\"{a}}s, and H.~Korkalainen, ``{Assessment of obstructive sleep
  apnea-related sleep fragmentation utilizing deep learning-based sleep staging
  from photoplethysmography},'' \emph{Sleep}, no. June, pp. 1--10, 2021.

\bibitem{Sutskever2014}
I.~Sutskever, O.~Vinyals, and Q.~V. Le, ``{Sequence to Sequence Learning with
  Neural Networks},'' \emph{Advances in Neural Information Processing Systems},
  vol.~4, no. January, pp. 3104--3112, 9 2014.

\bibitem{Quan1997}
S.~F. Quan, B.~V. Howard, C.~Iber, J.~P. Kiley, F.~J. Nieto, G.~T. O'Connor,
  D.~M. Rapoport, S.~Redline, J.~Robbins, J.~M. Samet, and P.~W. Wahl, ``{The
  Sleep Heart Health Study: design, rationale, and methods.}'' \emph{Sleep},
  vol.~20, no.~12, pp. 1077--85, 12 1997.

\bibitem{Zhang2018}
G.-Q. Zhang, L.~Cui, R.~Mueller, S.~Tao, M.~Kim, M.~Rueschman, S.~Mariani,
  D.~Mobley, and S.~Redline, ``{The National Sleep Research Resource: towards a
  sleep data commons},'' \emph{Journal of the American Medical Informatics
  Association}, vol.~25, no.~10, pp. 1351--1358, 10 2018.

\bibitem{Chen2015}
X.~Chen, R.~Wang, P.~Zee, P.~L. Lutsey, S.~Javaheri, C.~Alc{\'{a}}ntara, C.~L.
  Jackson, M.~A. Williams, and S.~Redline, ``{Racial/Ethnic Differences in
  Sleep Disturbances: The Multi-Ethnic Study of Atherosclerosis (MESA)},''
  \emph{SLEEP}, 6 2015.

\bibitem{Redline1995}
S.~Redline, P.~V. Tishler, T.~D. Tosteson, J.~Williamson, K.~Kump, I.~Browner,
  V.~Ferrette, and P.~Krejci, ``{The Familial Aggregation of Obstructive Sleep
  Apnea},'' \emph{American Journal of Respiratory and Critical Care Medicine},
  vol. 151, no. 3{\_}pt{\_}1, pp. 682--687, 3 1995.

\bibitem{Berry2017}
R.~B. Berry, R.~Brooks, C.~Gamaldo, S.~M. Harding, R.~M. Lloyd, S.~F. Quan,
  M.~T. Troester, and B.~V. Vaughn, ``{AASM Scoring Manual Updates for 2017
  (Version 2.4)},'' \emph{Journal of Clinical Sleep Medicine : JCSM : Official
  Publication of the American Academy of Sleep Medicine}, vol.~13, no.~5, p.
  665, 2017.

\bibitem{Hara2017}
S.~Hara, T.~Shimazaki, H.~Okuhata, H.~Nakamura, T.~Kawabata, K.~Cai, and
  T.~Takubo, ``{Parameter optimization of motion artifact canceling PPG-based
  heart rate sensor by means of cross validation},'' in \emph{2017 11th
  International Symposium on Medical Information and Communication Technology
  (ISMICT)}.\hskip 1em plus 0.5em minus 0.4em\relax IEEE, 2 2017, pp. 73--76.

\bibitem{Park2017}
C.~Park, H.~Shin, and B.~Lee, ``{Blockwise PPG Enhancement Based on
  Time-Variant Zero-Phase Harmonic Notch Filtering},'' \emph{Sensors}, vol.~17,
  no.~4, p. 860, 4 2017.

\bibitem{Pollreisz2019}
D.~Pollreisz and N.~TaheriNejad, ``{Detection and Removal of Motion Artifacts
  in PPG Signals},'' \emph{Mobile Networks and Applications}, pp. 1--11, 8
  2019.

\bibitem{Tanweer2017}
K.~T. Tanweer, S.~R. Hasan, and A.~M. Kamboh, ``{Motion artifact reduction from
  PPG signals during intense exercise using filtered X-LMS},'' in \emph{2017
  IEEE International Symposium on Circuits and Systems (ISCAS)}.\hskip 1em plus
  0.5em minus 0.4em\relax IEEE, 5 2017, pp. 1--4.

\bibitem{Temko2017}
A.~Temko, ``{Accurate Heart Rate Monitoring during Physical Exercises Using
  PPG},'' \emph{IEEE Transactions on Biomedical Engineering}, vol.~64, no.~9,
  pp. 2016--2024, 9 2017.

\bibitem{Aboy2005}
M.~Aboy, J.~McNames, T.~Thong, D.~Tsunami, M.~Ellenby, and B.~Goldstein, ``{An
  Automatic Beat Detection Algorithm for Pressure Signals},'' \emph{IEEE
  Transactions on Biomedical Engineering}, vol.~52, no.~10, pp. 1662--1670, 10
  2005.

\bibitem{Charlton2019}
P.~H. Charlton, J.~Mariscal~Harana, S.~Vennin, Y.~Li, P.~Chowienczyk, and
  J.~Alastruey, ``{Modeling arterial pulse waves in healthy aging: a database
  for in silico evaluation of hemodynamics and pulse wave indexes},''
  \emph{American Journal of Physiology-Heart and Circulatory Physiology}, vol.
  317, no.~5, pp. H1062--H1085, 11 2019.

\bibitem{Kotzen2021}
K.~Kotzen, P.~H. Charlton, A.~Landesberg, and J.~A. Behar, ``{Benchmarking
  Photoplethysmography Peak Detection Algorithms Using the Electrocardiogram
  Signal as a Reference},'' \emph{2021 Computing in Cardiology (CinC)}, pp.
  1--4, 9 2021.

\bibitem{Hamilton2002}
P.~Hamilton, ``{Open source ECG analysis},'' in \emph{Computers in
  Cardiology}.\hskip 1em plus 0.5em minus 0.4em\relax IEEE, 9 2002, pp.
  101--104.

\bibitem{Chocron2020}
A.~Chocron, R.~Efraim, F.~Mandel, M.~Rueschman, N.~Palmius, T.~Penzel,
  M.~Elbaz, and J.~A. Behar, ``{Digital biomarkers and artificial intelligence
  for mass diagnosis of atrial fibrillation in a population sample at risk of
  sleep disordered breathing},'' \emph{Physiological Measurement}, vol.~41,
  no.~10, 7 2020.

\bibitem{Oord2016}
A.~v.~d. Oord, S.~Dieleman, H.~Zen, K.~Simonyan, O.~Vinyals, A.~Graves,
  N.~Kalchbrenner, A.~Senior, and K.~Kavukcuoglu, ``{WaveNet: A Generative
  Model for Raw Audio},'' 9 2016.

\bibitem{Schneider2019}
S.~Schneider, A.~Baevski, R.~Collobert, and M.~Auli, ``{wav2vec: Unsupervised
  Pre-training for Speech Recognition},'' 4 2019.

\bibitem{Raghu2019}
M.~Raghu, C.~Zhang, J.~Kleinberg, and S.~Bengio, ``{Transfusion: Understanding
  Transfer Learning for Medical Imaging},'' \emph{Advances in Neural
  Information Processing Systems}, vol.~32, 2 2019.

\bibitem{Huy2021}
H.~Phan and K.~Mikkelsen, ``{Automatic Sleep Staging: Recent Development,
  Challenges, and Future Directions},'' pp. 1--28, 2021.

\bibitem{Lee2021}
Y.~J. Lee, J.~Y. Lee, J.~H. Cho, and J.~H. Choi, ``{Inter-rater reliability of
  sleep stage scoring: a meta-analysis},'' \emph{Journal of Clinical Sleep
  Medicine}, 7 2021.

\bibitem{Danker-Hopfe2009}
H.~Danker-Hopfe, P.~Anderer, J.~Zeitlhofer, M.~Boeck, H.~Dorn, G.~Gruber,
  E.~Heller, E.~Loretz, D.~Moser, S.~Parapatics, B.~Saletu, A.~Schmidt, and
  G.~Dorffner, ``{Interrater reliability for sleep scoring according to the
  Rechtschaffen {\&} Kales and the new AASM standard.}'' \emph{Journal of sleep
  research}, vol.~18, no.~1, pp. 74--84, 3 2009.

\bibitem{Elgendi2012}
M.~Elgendi, ``{On the analysis of fingertip photoplethysmogram signals},''
  \emph{Current cardiology reviews}, vol.~8, no.~1, pp. 14--25, 2012.

\bibitem{Li2018}
K.~Li, W.~Pan, Q.~Jiang, and G.-Z. Liu, ``{A Method to Detect Sleep Apnea based
  on Deep Neural Network and Hidden Markov Model using Single-Lead ECG
  signal},'' \emph{Neurocomputing}, vol. 294, 3 2018.

\bibitem{LeeChi2015}
T.~L. Lee-Chiong, \emph{{Sleep: A Comprehensive Handbook}}, L.-C. Teofilo~L,
  Ed.\hskip 1em plus 0.5em minus 0.4em\relax John Wiley {\&} Sons, 2005.

\end{thebibliography}

\newpage

\section*{Acknowledgment}
\textbf{The Multi-Ethnic Study of Atherosclerosis} (MESA) Sleep Ancillary study was funded by NIH-NHLBI Association of Sleep Disorders with Cardiovascular Health Across Ethnic Groups (RO1 HL098433). MESA is supported by NHLBI funded contracts HHSN268201500003I, N01-HC-95159, N01-HC-95160, N01-HC-95161, N01-HC-95162, N01-HC-95163, N01-HC-95164, N01-HC-95165, N01-HC-95166, N01-HC-95167, N01-HC-95168 and N01-HC-95169 from the National Heart, Lung, and Blood Institute, and by cooperative agreements UL1-TR-000040, UL1-TR-001079, and UL1-TR-001420 funded by NCATS. The National Sleep Research Resource was supported by the National Heart, Lung, and Blood Institute (R24 HL114473, 75N92019R002).\\
\textbf{The Cleveland Family Study} (CFS) database is only available for non-commercial use. The Cleveland Family Study (CFS) was supported by grants from the National Institutes of Health (HL46380, M01 RR00080-39, T32-HL07567, RO1-46380). The National Sleep Research Resource was supported by the National Heart, Lung, and Blood Institute (R24 HL114473, 75N92019R002).\\
\textbf{The Sleep Heart Health Study} (SHHS) was supported by National Heart, Lung, and Blood Institute cooperative agreements U01HL53916 (University of California, Davis), U01HL53931 (New York University), U01HL53934 (University of Minnesota), U01HL53937 and U01HL64360 (Johns Hopkins University), U01HL53938 (University of Arizona), U01HL53940 (University of Washington), U01HL53941 (Boston University), and U01HL63463 (Case Western Reserve University). The National Sleep Research Resource was supported by the National Heart, Lung, and Blood Institute (R24 HL114473, 75N92019R002).

\onecolumn 

\appendices
\section{List of patients in MESA-test}\label{secA1}
0001, 0021, 0033, 0052, 0077, 0081, 0101, 0111, 0225, 0310, 0314, 0402, 0416, 0445, 0465, 0483, 0505, 0554, 0572, 0587, 0601, 0620, 0648, 0702, 0764, 0771, 0792, 0797, 0800, 0807, 0860, 0892, 0902, 0904, 0921, 1033, 1080, 1121, 1140, 1148, 1161, 1164, 1219, 1224, 1271, 1324, 1356, 1391, 1463, 1483, 1497, 1528, 1531, 1539, 1672, 1693, 1704, 1874, 1876, 1900, 1914, 2039, 2049, 2096, 2100, 2109, 2169, 2172, 2183, 2208, 2239, 2243, 2260, 2269, 2317, 2362, 2388, 2470, 2472, 2488, 2527, 2556, 2602, 2608, 2613, 2677, 2680, 2685, 2727, 2729, 2802, 2811, 2828, 2877, 2881, 2932, 2934, 2993, 2999, 3044, 3066, 3068, 3111, 3121, 3153, 3275, 3298, 3324, 3369, 3492, 3543, 3554, 3557, 3561, 3684, 3689, 3777, 3793, 3801, 3815, 3839, 3886, 3997, 4110, 4137, 4171, 4227, 4285, 4332, 4406, 4460, 4462, 4497, 4501, 4552, 4577, 4649, 4650, 4667, 4732, 4794, 4888, 4892, 4895, 4912, 4918, 4998, 5006, 5075, 5077, 5148, 5169, 5203, 5232, 5243, 5287, 5316, 5357, 5366, 5395, 5397, 5457, 5472, 5479, 5496, 5532, 5568, 5580, 5659, 5692, 5706, 5737, 5754, 5805, 5838, 5847, 5890, 5909, 5957, 5983, 6015, 6039, 6047, 6123, 6224, 6263, 6266, 6281, 6291, 6482, 6491, 6502, 6516, 6566, 6567, 6583, 6619, 6629, 6646, 6680, 6722, 6730, 6741, 6788

\section{FE Model Diagrams}\label{secA2}
\begin{table*}[h!]
\label{tab:summary_features_network}
\caption{Description of the network used to perform Sleep Staging using Feature Engineering.}
\begin{center}
\addtolength{\leftskip} {-6cm} 
\addtolength{\rightskip}{-6cm}
\begin{tabular*}{\textwidth}{p{0.2\textwidth}p{0.25\textwidth}p{0.5\textwidth}}
\hline
Shape & Layer & Description \\
\hline \hline
$1200 \times n_{x}$ & Input & $n_{x}$ is the number of features \\
& & \\
$1200 \times 512$ & Dense & \\
$1200 \times 256$ & Dense & \\
$1200 \times 128$ & Dense & \\
$1200 \times 16$ & Dense & \\
$1200 \times n_{e}$ & Dense & $n_{e}$ is the embedding size \\
& & \\
$1200 \times 2n_{h}$ & LSTM & $n_{h}$ is the number of hidden units \\
$1200 \times 2n_{h}$ & LSTM & $n_{h}$ is the number of hidden units \\
& & \\
 & Dropout & \\
$1200 \times 256$ & Dense & \\
 & Dropout & \\
$1200 \times 128$ & Dense & \\
 & Dropout & \\
$1200 \times 64$ & Dense & \\
$1200 \times 4$ & Dense & Softmax Activation \\

\hline
\end{tabular*}
\end{center}
\end{table*}

\section{DTS Model Diagrams}\label{secA3}
\begin{table*}[h!]
\label{tab:summary_IPR_network}
\caption{Description of the network used to perform Sleep Staging using IPR time series.}
\begin{center}
\addtolength{\leftskip} {-6cm} 
\addtolength{\rightskip}{-6cm}
\begin{tabular*}{\textwidth}{p{0.2\textwidth}p{0.2\textwidth}p{0.1\textwidth}p{0.1\textwidth}p{0.3\textwidth}}
\hline
Shape & Layer & & & Description \\
\hline \hline
$1200 \times 256 \times 1$ & Input & & & 10 hours 2Hz \\
 ~ & ~ & ~ & ~ & ~ \\
$1200 \times 256 \times F$ & 1D-Conv & $\|$ $- \times 2$ & $\|$ & $i = [1,2,3]$ \\
$1200 \times 128/i \times F$ & MaxPool & & $\|$ - X 3 & F = [16, 32, 64] \\
$1200 \times 128/i \times F$ & Residual & & $\|$ & \\
$1200 \times 2048$ & Flatten & & & \\
$1200 \times 128$ & Dense & & & \\
$1 \times 1200 \times 128$ & Reshape & & & \\
$1 \times 1200 \times 128$ & Dilated 1D-Conv & $\|$ -X 6 & $\|$ & kernel = 7 \\
$1 \times 1200 \times 128$ & Residual & & $\|$ - X 2 & dilation = [1,2,4,8,16,32] \\
$1 \times 1200 \times 128$ & Dropout & & $\|$ & \\
 & & & & \\
$1 \times 1200 \times 4$ & 1D-Conv & & & Softmax Activation \\
\hline
\end{tabular*}
\end{center}
\end{table*}

\end{document}